\documentclass[10pt]{article}

\usepackage{amsthm,amsmath}

\usepackage[utf8]{inputenc} %unicode support

\usepackage{nameref,hyperref}
\usepackage{graphicx}
\usepackage{xcolor}
\usepackage{multirow}
\usepackage{bold-extra}
\usepackage{tabularx}
\usepackage[
  backend=biber,
  style=numeric,
  sorting=none
]{biblatex}
\usepackage[a4paper,
  twoside=false,
  vmargin=1cm,
  hmargin=1in,
  centering,
  columnsep=2pc,
  heightrounded,
  includeheadfoot,
]{geometry}

\usepackage{caption}
\AtBeginEnvironment{table}{\small}
\AtBeginEnvironment{figure}{\small}

\begin{document}
\vspace*{0.35in}

\begin{flushleft}
{\Large
\textbf\newline{Clinical Data Goes MEDS? Let's OWL make sense of it}
}
\smallskip

% authors go here:
Alberto Marfoglia\textsuperscript{1,2}, 
Jong Ho Jhee\textsuperscript{2},
Adrien Coulet\textsuperscript{2,$\star$}\\

\bigskip

\text{\textsuperscript{1}} Dept. of Computer Science and Engineering - DISI, University of Bologna, Bologna, Italy
\\
\text{\textsuperscript{2}} Inria, Inserm, Université Paris Cité, HeKA, UMR 1346, Paris, France
\\

\bigskip
%\textsuperscript{$\diamond$} These authors contributed equally to this work.\\
\textsuperscript{$\star$} corresponding author: \texttt{adrien.coulet@inria.fr}

\end{flushleft}

\bigskip
\bigskip

\abstract{
The application of machine learning on healthcare data is often hindered by the lack of standardized and semantically explicit representation, leading to limited interoperability and reproducibility across datasets and experiments.
The Medical Event Data Standard (MEDS) addresses these issues by introducing a minimal, event-centric data model designed for reproducible machine-learning workflows from health data. However, MEDS is defined as a data-format specification and does not natively provide integration with the Semantic Web ecosystem.
In this article, we introduce MEDS-OWL, a lightweight OWL ontology that provides formal concepts and relations to represent MEDS datasets as RDF graphs. 
Additionally, we implemented \texttt{meds2rdf}, a Python conversion library that transforms MEDS events into RDF graphs, ensuring conformance with the ontology.
We evaluate the proposed approach on two datasets: a synthetic clinical cohort describing care pathways for ruptured intracranial aneurysms, and a real-world subset of MIMIC-IV. To assess semantic consistency, we performed a SHACL validation against the resulting knowledge graphs.
The first release of MEDS-OWL comprises 13 classes, 10 object properties, 20 data properties, and 24 OWL axioms. Combined with \texttt{meds2rdf}, it enables data transformation into FAIR-aligned datasets, provenance-aware publishing, and interoperability of event-based clinical data. By bridging MEDS with the Semantic Web, this work contributes a reusable semantic layer for event-based clinical data and establishes a robust foundation for subsequent graph-based analytics.}

%\maketitle
\bigskip
\bigskip

\section{Introduction}
Machine learning (ML) in the healthcare domain is hindered by the lack of a standardized representation for structured Electronic Health Record (EHR) data, resulting in dataset-specific preprocessing practices, limited reproducibility, and poor model portability across datasets. The Medical Event Data Standard (MEDS)~\cite{bert_arnrich_medical_2024, mcdermott_meds_2024} addresses this gap by introducing a unified, event-centric schema for longitudinal health data that is explicitly designed to support reproducible and efficient ML workflows. MEDS encodes each event as a minimal tuple consisting of a subject identifier, a code, and an optional temporal and value attributes. This minimal and pragmatically oriented design makes MEDS particularly suitable for benchmarking, multi-center studies, and the development of foundational models. For these reasons, MEDS is being adopted in various ML projects as a common schema for EHR data and has been utilized in a transformer-based model for predicting future events in patient healthcare  trajectories~\cite{renc_foundation_2025, oufattole_meds-tab_2024}.

In its native form, MEDS is a data-format specification and does not provide ontology-level semantics. The Semantic Web paradigm offers a complementary approach to address this limitation~\cite{hitzler_foundations_2009, carbonaro_connected_2023, marfoglia_knowledge_2026}. First, ontologies expressed in Resource Description Framework (RDF) and Web Ontology Language (OWL) provide an interoperable, machine-interpretable lingua franca for representing entities and their relationships.  Second, they support semantic querying and formal reasoning. Third, they support the construction of knowledge graphs (KGs) that facilitate the explicit linkage of dataset elements to external terminologies such as LOINC, SNOMED CT, and ATC~\cite{hogan_knowledge_2021}.
Projecting MEDS datasets into RDF would thus enable their semantic integration and validation, while also creating opportunities to explore graph-based ML on event-oriented datasets. Indeed, patient representations can be learned from graph representations with several advantages. KG embeddings map graph nodes and edges into a continuous vector space~\cite{hamilton_representation_2018}, enabling the representation of complex relational structures in a form amenable to computational analysis. Such approaches have been successfully applied to tasks including link prediction, node classification, and clustering~\cite{wang_knowledge_2017}. 
In the clinical domain, relational models such as Relational Graph Convolutional Networks (RGCNs) have been shown to outperform tabular baselines when the underlying schema aligns with task-specific requirements; for instance, compact and patient-centered representations for outcome prediction \cite{jhee_predicting_2025}. These observations motivate the present work. 

While several data models and ontologies, such as FHIR~\cite{lehne_use_2019}, OMOP CDM~\cite{xiao_fhir-ontop-omop_2022}, and Phenopackets~\cite{kaliyaperumal_phenopackets_2022}, have been proposed to harmonize clinical data, RDF transformations of these models have seen limited adoption~\cite{xiao_fhir-ontop-omop_2022, marfoglia_towards_2025}, in part due to loosely specified semantics or incomplete coverage of the clinical domain.
% OLD: Ontologies such as SPHN~\cite{toure_fairification_2023} and CARE-SM~\cite{kaliyaperumal_semantic_2022} illustrate that patient-centric KG representations can facilitate interoperability, provenance modeling, and temporal reasoning. 
% NEW:
Patient-centric RDF-based representations such as SPHN~\cite{toure_fairification_2023} and CARE-SM~\cite{kaliyaperumal_semantic_2022} illustrate how KGs can facilitate interoperability, provenance modeling, and temporal reasoning.
However, these models are often tightly coupled to specific use cases or semantic frameworks.

In this study, we propose to bridge MEDS with the Semantic Web formalism. We report on the development of two complementary components: MEDS-OWL, an OWL ontology that captures the core concepts and relations of MEDS \cite{marfoglia_teamhekameds-ontology_2025}, and \texttt{meds2rdf}, a Python library that transforms MEDS-compliant datasets into RDF graphs in accordance with the ontology \cite{marfoglia_teamhekameds2rdf_2025}. Together, these components enable interoperable, reproducible, and FAIR-aligned representations of event-based clinical data.

The remainder of this article is organized as follows. Section~\ref{sec:methods} describes the methodology adopted for the development of the MEDS-OWL ontology and the accompanying conversion library. Section~\ref{sec:results} presents the resulting ontology and the \texttt{meds2rdf} library in action. Finally, Section~\ref{sec:discussion} discusses the results and outlines directions for future work.

\section{Methodology}\label{sec:methods}
The MEDS-OWL ontology was developed following the principled and iterative methodology described by Noy and McGuinness in “Ontology Development 101”~\cite{noy_ontology_2001}, using the Protégé ontology editor~\cite{musen_protege_2015}. The scope of the ontology encompasses the representation of EHR and claims-style medical events as defined by the MEDS conceptual model, together with the dataset- and experiment-level metadata required to support reproducible ML workflows. As part of this process, we formulated a set of competency questions to guide the modeling choices and to ensure that the ontology adequately supports the intended use cases. An example of such a competency question is: \emph{“Given a medical event, which subject does it pertain to, when did it occur, and which canonical code and value modalities are associated with it?”}. These competency questions informed the identification and formalization of the core MEDS classes, namely \texttt{Subject}, \texttt{Event}, \texttt{Code}, \texttt{DatasetMetadata}, \texttt{SubjectSplit}, \texttt{SubjectLabel}, and \texttt{ValueModality}. A detailed description of these %six%
Seven core classes, together with illustrative examples, are provided in Table~\ref{tab:meds-owl-concepts}.

\begin{table}
\centering
\caption{Core MEDS-OWL classes and their interpretation.}
\label{tab:meds-owl-concepts}
\begin{tabular}{p{3cm} p{7.6cm} p{4cm}}
\hline
\textbf{Class} & \textbf{Description} & \textbf{Example} \\
\hline
\texttt{Subject} &
Primary entity under observation, typically corresponding to a patient, admission, or longitudinal unit to which events are attributed. &
Patient \#10234 \\
\texttt{Event} &
Atomic clinical observation representing a single MEDS row, characterized by a subject, a coded concept, and optional temporal and value attributes. &
Administration of nimodipine at admission \\
\texttt{Code} &
An entity used to type events, which includes canonical identifiers, descriptions, and hierarchical relations, with optional linkage to external terminologies. &
ATC:C08CA06 (Nimodipine) \\
\texttt{DatasetMetadata} &
Dataset-level description and provenance information, capturing identity, versioning, creation time, and ETL-related metadata. &
X dataset, in MEDS format, v1.0 \\
\texttt{SubjectSplit} &
A grouping of subjects in several disjoint sets aimed for an ML experiment. &
Split for experiment 5 into a train, eval, and test sets\\
\texttt{SubjectLabel} &
A supervised learning instance defined by a subject identifier, an as-of prediction time (the latest time up to which data may be used), and exactly one observed ground truth label value \(y\). &
Subject \#10234, prediction time = discharge date, categorical label = rehabilitation\\
% NEW:
\texttt{ValueModality} & Generic, extensible entity that links an \texttt{Event} to additional value modalities (for example, images, waveform recordings, audio clips, or other non-tabular payloads). & Chest X-ray image (DICOM series URI)\\
\hline
\end{tabular}
\end{table}

To clarify how these ontology classes relate to the original MEDS data model, 
Table~\ref{tab:schema-to-ontology} summarizes the correspondence between the five primary MEDS schema components and the MEDS-OWL classes derived from them.
In MEDS, each schema is defined as an Apache Arrow schema, specifying the required columns and data types for MEDS data files. 
\begin{table}
\centering
\caption{Mapping from MEDS schema components to MEDS-OWL classes. See Table~\ref{tab:meds-owl-concepts} for class definitions and examples.}
\label{tab:schema-to-ontology}
\begin{tabular}{p{3.4cm} p{5.2cm} p{6cm}}
\hline
\textbf{MEDS schema} & \textbf{Brief description} & \textbf{Mapped MEDS-OWL class(es)} \\
\hline
DataSchema &
Core medical event table describing sequences of subject observations, where each row corresponds to a single clinical event. &
\texttt{meds:Event} as the primary representation of each MEDS row.  
\texttt{meds:ValueModality} is used to model additional or non-tabular payloads associated with an event (e.g., images, waveforms, signals).  
See Table~\ref{tab:events} for row-to-triple mappings. \\[4pt]

DatasetMetadataSchema &
Dataset-level metadata captures dataset identity, versioning, creation time, licensing, and ETL provenance information. &
\texttt{meds:DatasetMetadata}, aligned with \texttt{dcterms} and \texttt{dcat}, and linked to ETL activities via \texttt{prov:wasGeneratedBy}.  
See Table~\ref{tab:dataset}. \\[4pt]

CodeMetadataSchema &
Metadata describing the controlled vocabularies and codes used to type medical events, including canonical identifiers and hierarchical relations. &
\texttt{meds:Code}, with \texttt{meds:codeString} as a canonical identifier and optional \texttt{meds:parentCode} relations to represent code hierarchies.  
See Table~\ref{tab:codes}. \\[4pt]

SubjectSplitSchema &
Specification of how subjects are assigned to mutually exclusive subsets (e.g., training, tuning, held-out) within a given machine learning experiment. &
\texttt{meds:SubjectSplit}, with subjects linked via \texttt{meds:assignedSplit}.  
Multiple splits may coexist across experiments, while each subject belongs to exactly one subset per split.  
See Table~\ref{tab:labels}. \\[4pt]

LabelSchema &
Definition of supervised learning labels associated with a subject at a specific prediction time, including the observed ground truth value. &
\texttt{meds:SubjectLabel}, capturing the prediction cutoff time (\texttt{meds:predictionTime}) and exactly one label value using datatype-specific properties.  
See Table~\ref{tab:labels}. \\[4pt]

(subject identifier) &
Subject identifiers appear across multiple MEDS schemas and provide the logical join key between events, splits, and labels. &
\texttt{meds:Subject}, representing the canonical subject entity via \texttt{meds:subjectId}, and serving as the hub linking \texttt{Event}, \texttt{SubjectLabel}, and \texttt{SubjectSplit}. \\

\hline
\end{tabular}
\end{table}

The OWL classes and properties we developed closely follow the MEDS data model, while reusing as much as possible established external ontologies to maximize interoperability. Specifically, we reused the Data Catalogue Vocabulary (DCAT), DCTERMS, and PROV ontology (PROV-O) to represent datasets, ETL (Extraction, Transformation, Loading) processes, and provenance metadata, respectively, thereby ensuring alignment with widely adopted ontologies. Finally, for MEDS fields specified as optional (nullable), such as time, the ontology captures this constraint through upper-bound cardinality restrictions (\texttt{owl:maxCardinality 1}) without corresponding minimum bound or existential constraints, thus permitting zero or one value, while forbidding multiple values.

\begin{figure}
  \centering
  \includegraphics[width=0.9\textwidth]{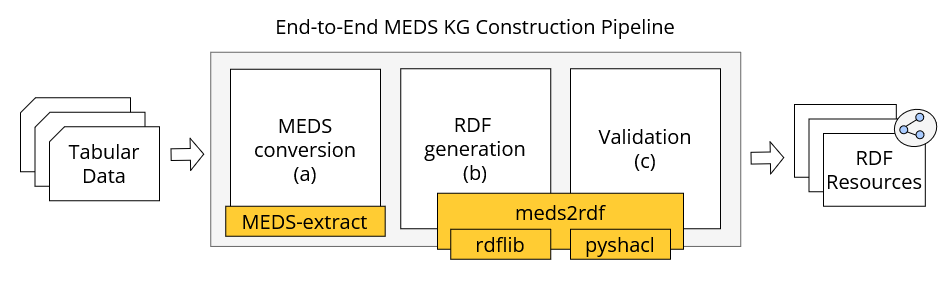}
  \caption{Overview of the end-to-end MEDS KG construction pipeline, illustrating (a) normalization and extraction of tabular data into MEDS event tuples (optional, if data are already in MEDS), (b) transformation into RDF KGs compliant with the MEDS-OWL ontology, and (c) SHACL-based validation and serialization into standard RDF formats.}
  \label{fig:meds-to-rdf-etl}
\end{figure}
We developed a Python library named \texttt{meds2rdf}~\cite{marfoglia_teamhekameds2rdf_2025} to implement the mapping rules necessary to transform a MEDS dataset into an RDF graph that follows the terms of the MEDS-OWL ontology.
Figure~\ref{fig:meds-to-rdf-etl} presents the end-to-end conversion workflow implemented leveraging \texttt{meds2rdf}. The pipeline consists of three main stages. 
(a) First, source EHR data are ingested by the MEDS-extract component~\cite{mcdermott_meds_2024}, which applies a declarative conversion template to map dataset-specific features into MEDS event tuples. Note that this first step is optional, and only necessary when healthcare data are not already available in MEDS. 
% [OLD]: (b) In the subsequent RDF generation stage, MEDS-formatted data are transformed into KG conforming to the MEDS-OWL ontology using the \texttt{meds2rdf} library. To implement this library, we established manually mapping tables that systematically relate elements of the MEDS schema to the corresponding classes and properties in MEDS-OWL. These tables constitute a central result of this study, as they provide an implementation-independent specification of how tabular MEDS datasets are transformed into RDF while preserving their intended semantics.
(b) In the subsequent RDF generation stage, MEDS-formatted data are transformed into a KG conforming to the MEDS-OWL ontology using the \texttt{meds2rdf} library. The mapping module of \texttt{meds2rdf} implements a deterministic, table-driven transformation from MEDS tuples to RDF. It exploits manually curated mapping tables that relate elements of the MEDS schema to corresponding classes and properties in MEDS-OWL.
These tables constitute a central contribution of this work, as they provide an implementation-independent and ontology-aligned specification that formalizes the MEDS-to-RDF mapping. Moreover, in this phase, \texttt{meds2rdf} focuses on syntactic correctness and local mapping guarantees, such as required-field checks and type normalization. The output of this stage is an in-memory RDF graph constructed using the \texttt{rdflib.Graph} data structure.
% [OLD]: (c) Next, the resulting RDF graph is validated against a SHACL schema that formally encodes MEDS constraints, supporting data quality assessment and semantic correctness. Finally, validated graphs can be serialized in standard RDF formats, including Turtle, RDF/XML, and N-Triples.
(c) Next, the generated graph is validated against a SHACL schema using the \texttt{pySHACL} validator, which is integrated within \texttt{meds2rdf}. The schema declaratively encodes the semantic constraints of MEDS-OWL, and the validation process ensures the enforcement of both global and cross-entity invariants. These include mutually exclusive property patterns, cardinality constraints, and referential consistency across entities. This approach allows semantic validation to be performed before graph materialization. Finally, RDF graphs that successfully pass SHACL validation may be serialized as an optional final step in standard RDF formats, including Turtle, RDF/XML, and N-Triples.

Integrating SHACL validation within \texttt{meds2rdf} enables the ontology and its constraints to evolve independently of the mapping logic. This design choice is crucial given that MEDS-OWL is currently at its first public version and is expected to undergo further revisions in alignment with the evolving MEDS specification. By leveraging the declarative nature of SHACL, future updates to the ontology or constraint definitions can be accommodated by modifying the SHACL schema alone, without requiring changes to the conversion code.
\section{Results and examples of usage}\label{sec:results}
\begin{figure}
  \centering
  \includegraphics[width=1.0\textwidth]{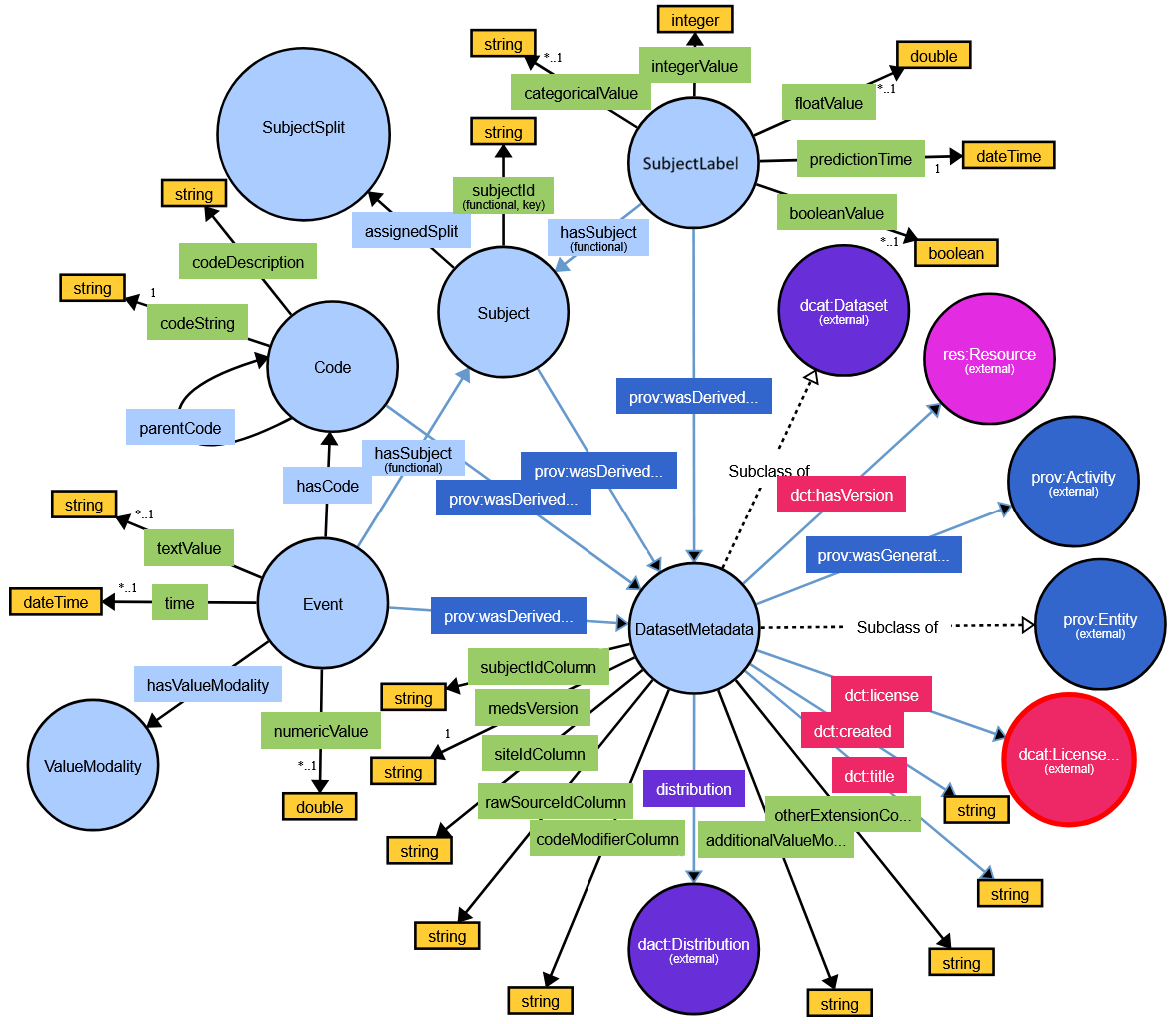}
\caption{Overview of the MEDS-OWL ontology. Azure nodes and edges denote locally defined classes and object properties, respectively. Black edges with green labels are datatype properties. Nodes and edges colored differently indicate imported entities from either the Data Catalogue Vocabulary (\texttt{dcat} prefix), Dublin Core Terms (\texttt{dct} prefix) or PROV-O (prefix \texttt{prov}). Dotted edges represent \texttt{subClassOf} predicate. The figure is drawn with WebVOWL~\cite{lohmann2014webvowl}.}
  \label{fig:ontology-diagram}
\end{figure}

%\textbf{MEDS-OWL} operationalizes MEDS within the Semantic Web paradigm, enabling provenance-aware publishing, semantic validation, and linkage to external biomedical ontologies.
\textbf{MEDS-OWL} in its first release comprehends  13 classes, 10 object properties, 20 data properties, and 24 OWL axioms, and it is accompanied by example RDF instances and a SHACL validation suite. MEDS-OWL is illustrated in Figure~\ref{fig:ontology-diagram} and its documentation is available on Zenodo and GitHub~\cite{marfoglia_teamhekameds-ontology_2025}.

%This process produced a set of mapping tables that systematically relate elements of the MEDS schema to the corresponding classes and properties in MEDS-OWL. These tables constitute a central result of this study, as they provide an implementation-independent specification of how tabular MEDS datasets are transformed into RDF while preserving their intended semantics.

\textbf{Mappings} between MEDS core data schema elements and MEDS-OWL are reported in Tables~\ref{tab:events}-\ref{tab:labels}. Table~\ref{tab:events} reports those to the \texttt{meds:Event} class. In MEDS-OWL, events are required to be associated with exactly one subject and exactly one code, while temporal and value-related attributes are optional and for this reason constrained by cardinality restrictions.  Table~\ref{tab:codes} reports mappings to the \texttt{meds:Code} class. Code metadata is modeled explicitly as individuals of type \texttt{meds:Code}, which enables reusability, hierarchical organization, and linkage to external terminologies. Dataset-level provenance and conversion metadata are represented by the \texttt{meds:DatasetMetadata} class, which is aligned with both \texttt{dcat:Dataset} and \texttt{prov:Entity}. Cardinality constraints are enforced at the OWL level. The corresponding mappings are shown in Table~\ref{tab:dataset}. Finally, Table~\ref{tab:labels} reports the mapping for dataset splits and supervised learning labels. Dataset partitions are modeled as predefined individuals of class \texttt{meds:SubjectSplit}, while label samples are represented as instances of \texttt{meds:SubjectLabel} with exactly one label value.
The \textbf{meds2rdf library} that implements the mapping rules is available on Zenodo and GitHub \cite{marfoglia_teamhekameds2rdf_2025}.
\begin{table}
\centering
\caption{Mapping of the elements of the MEDS DataSchema tables to \texttt{meds:Event} and \texttt{meds:Subject} classes, with descriptions. For each tuple (or row) of DataSchema, a list of associated RDF triples is generated. Card. stands for the cardinality of the relationship.}
\label{tab:events}
\begin{tabular}{l p{4cm} >{\raggedright}p{7cm} c}
\hline
\textbf{MEDS element} & \textbf{Description} & \textbf{RDF triple} & \textbf{Card.} \\
\hline
event\_row e & The row representing an event e of the subject s. & \texttt{?e a meds:Event.}  & 1 \\
subject\_id id& The identifier of the subject s (typically the patient). & \texttt{?s a meds:Subject. \break ?s meds:subjectId ?id}\verb|^^|\texttt{xsd:string.} \break \texttt{?e meds:hasSubject ?s.} & 1 \\
time t& The time of the measurement, nullable for static measurements. & \texttt{?e meds:time ?t}\verb|^^|\texttt{xsd:dateTime.} & 0..1 \\
code c & The categorical descriptor of the measurement (e.g., lab test, diagnosis code). & \texttt{?c a meds:Code. \break ?e meds:hasCode ?c.} & 1 \\
code\_string cs& The string representation of the code. & \texttt{?e meds:codeString ?cs.}\verb|^^|\texttt{xsd:string.} & 1 \\
numeric\_value nv & A numeric value associated with the measurement (e.g., lab test result), nullable. & \texttt{?e meds:numericValue ?nv}\verb|^^|\texttt{xsd:double.} & 0..1 \\
text\_value tv & A text value associated with the measurement (e.g., clinical note, text-based test result), nullable. & \texttt{?e meds:textValue ?tv}\verb|^^|\texttt{xsd:string.} & 0..1 \\
\hline
\end{tabular}
\end{table}
\begin{table}
\centering
\caption{Mapping of the elements of the MEDS CodeMetadataSchema table to \texttt{meds:Code} class. For each tuple (or row) of CodeMetadataSchema, a list of associated RDF triples is generated. } 
\label{tab:codes}
\begin{tabular}{l p{4cm} >{\raggedright}p{7cm} c}
\hline
\textbf{MEDS element} & \textbf{Description} & \textbf{RDF triple} & \textbf{Card.} \\
\hline
code\_row c & The row representing a code c in the dataset. & \texttt{?c a meds:Code.} & 1 \\
code\_string cs & The categorical descriptor of a possible measurement; links to the DataSchema code column. & \texttt{?c meds:codeString ?cs}\verb|^^|\texttt{xsd:string.} & 1 \\
description d& A human-readable description of what the code represents. & \texttt{?c meds:codeDescription ?d}\verb|^^|\texttt{xsd:string.} & 0..1 \\
parent\_codes p & One or more higher-level or parent codes in an ontological hierarchy, possibly linking to other codes in MEDS or external vocabularies (e.g., OMOP CDM). & \texttt{?p a meds:Code. \break ?c meds:parentCode ?p.} & 0..* \\
\hline
\end{tabular}
\end{table}
\begin{table}
\centering
\caption{Mapping metadata of a transformed MEDS dataset (DatasetMetadataSchema) to the \texttt{meds:DatasetMetdata} class. For each dataset generated, a list of associated ``meta'' RDF triples is generated.}
\label{tab:dataset}
\begin{tabular}{l p{4cm} >{\raggedright}p{7cm} c}
\hline
\textbf{MEDS element} & \textbf{Description} & \textbf{RDF triple} & \textbf{Card.} \\
\hline
dataset ds & The dataset converted to RDF. & \texttt{?ds a meds:DatasetMetadata.%\break rdfs:subClassOf prov:Entity, dcat:Dataset.
} & 1 \\
dataset\_name n & The name of the dataset ds. & \texttt{?ds dct:title ?n}\verb|^^|\texttt{xsd:string.} & 1 \\
dataset\_version dsv& The version of the dataset. & \texttt{?ds dct:hasVersion ?dsv.} & 0..1 \\
meds\_version mv& The version of the MEDS schema. & \texttt{?ds meds:medsVersion ?mv}\verb|^^|\texttt{xsd:string.} & 1 \\
created\_at & ISO 8601 creation date and time. & \texttt{?ds dct:created ?time}\verb|^^|\texttt{xsd:dateTime.} & 1 \\
license & License under which the dataset is released. & \texttt{?li a dct:LicenseDocument. \break ?ds dct:license ?li.} & 0..1 \\
location\_uri & URI(s) where the dataset is hosted. & \texttt{?ds dcat:distribution ?dist. \break ?dist dcat:downloadURL ?dw.} & 0..* \\
description\_uri & URI(s) referencing a detailed description of the dataset. & \texttt{?dist dcat:accessURL ?au.} & 0..* \\
etl\_name etln& Name of the ETL process etl that generated the dataset. & \texttt{?etl a prov:Activity. \break ?etl rdfs:label ?etln. \break ?ds prov:wasGeneratedBy ?etl.} & 0..1 \\
etl\_version etlv& Version of the ETL process. & 
\texttt{?etl dct:hasVersion ?etlv.} & 0..1 \\
% raw\_source\_id\_columns & Columns containing identifiers for the raw source data. & \texttt{?d meds:rawSourceIdColumns ?v} & 0..* \\
% code\_modifier\_columns & Columns that modify core codes. & \texttt{?d meds:codeModifierColumns ?v} & 0..* \\
% site\_id\_columns & Columns containing site-of-care identifiers. & \texttt{?d meds:siteIdColumns ?v} & 0..* \\
% other\_extension\_columns & Other non-core columns not described above. & \texttt{?d meds:otherExtensionColumns ?v} & 0..* \\
\hline
\end{tabular}
\end{table}
\begin{table}
\centering
\caption{Mapping of the elements of MEDS SubjectSplitSchema and LabelSchema tables to \texttt{meds:SubjectSplit} and \texttt{meds:SubjectLabel} classes, respectively. For each tuple (or row) of LabelSchema, a list of associated RDF triples is generated.
}
\label{tab:labels}
\begin{tabular}{l >{\raggedright}p{5cm} >{\raggedright}p{6cm} c}
\hline
\textbf{MEDS element} & \textbf{Description} & \textbf{RDF triple} & \textbf{Card.} \\
\hline
split & The name of the subset a subject is assigned to, following a dataset split (e.g., train, validation, test). & \texttt{?set a meds:SubjectSplit. \break ?s a meds:Subject. \break ?s meds:assignedSplit ?set.} & 0..* \\

label\_row & A prediction label (sample) for a subject in a supervised learning task. This is the ground truth label extracted from the dataset according to the task specification. & \texttt{?l a meds:SubjectLabel. \break ?l meds:hasSubject ?s.} & 0..* \\

prediction\_time & The timestamp representing the endpoint of the data that can be used to predict the label for the given sample. This is not necessarily the time of the event itself nor the deployment prediction time. It defines the "as-of" cutoff for input data. & \texttt{?l meds:predictionTime ?pt}\verb|^^|\texttt{xsd:dateTime.} & 1 \\

label\_value & The value for the label. It can be one of the following types depending on the task: binary tasks, ordinal or continuous numeric tasks, and categorical tasks. 
&
\texttt{?l meds:booleanValue ?bv}\verb|^^|\texttt{xsd:boolean.} \break
\texttt{?l meds:integerValue ?iv}\verb|^^|\texttt{xsd:integer.} \break
\texttt{?l meds:floatValue ?fv}\verb|^^|\texttt{xsd:double.} \break
\texttt{?l meds:categoricalValue ?cv}\verb|^^|\texttt{xsd:string.} & 0..1 \\
\hline
\end{tabular}
\end{table}

\textbf{Neurovasc example of usage}. To evaluate \texttt{meds2rdf}, we first employed a synthetic EHR dataset, here named \textit{neurovasc}, that describes care pathways of 10,000 patients hospitalized for a ruptured intracranial aneurysm~\cite{jhee_predicting_2025}. 
\textit{neurovasc} is an open dataset generated based on real-world EHR from Nantes University Hospital. Its patients are associated with 30 clinical features, including 8 time-stamped events representing the interval from hospital admission to their first occurrence, and 22 non-temporal demographic or historical features. Patient outcomes are categorized in one of three values: Back Home, Rehabilitation, or Death, depending on the patient's status at the end of the hospital stay. The dataset preserves realistic correlations and event transitions, making it suitable for testing ontology-driven data representation of clinical pathways and basic analytics.

\begin{figure}
  \centering
  \includegraphics[width=1.0\textwidth]{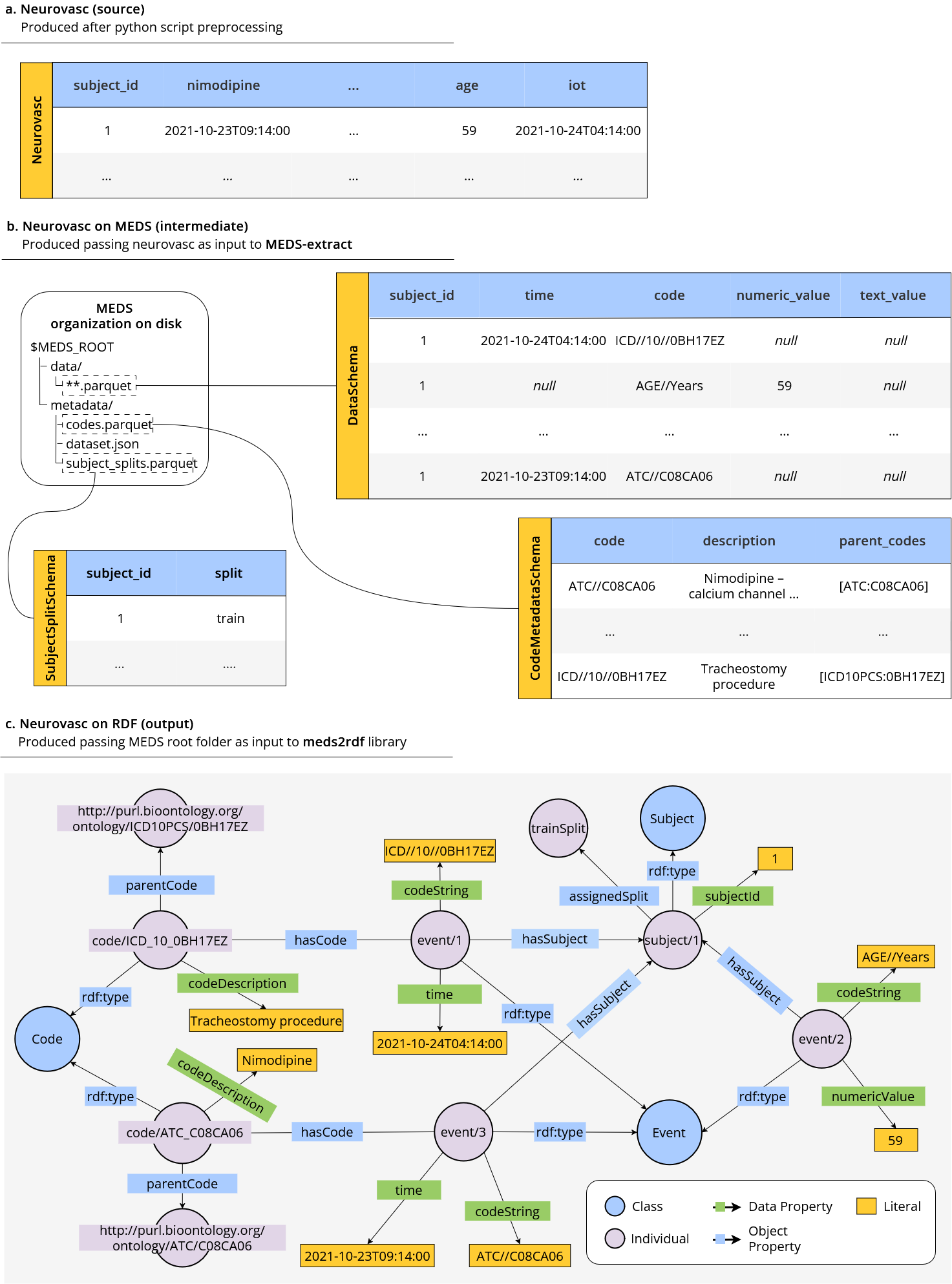}
  \caption{MEDS-OWL conversion workflow for the \textit{neurovasc} dataset. (a) Initial preprocessing of three patient features: age, nimodipine (drug administration), and iot (tracheostomy procedure). (b) Conversion to MEDS tabular format via MEDS-extract, with each feature as a distinct event. (c) Transformation of MEDS output into RDF graphs using \texttt{meds2rdf}.}
  \label{fig:etl-conversion}
\end{figure}

The conversion of the \textit{neurovasc} dataset illustrates the practical utility of MEDS-OWL. As shown in Figure~\ref{fig:etl-conversion}, the workflow depicts the output of each conversion stage, beginning with a simplified representation of the original \textit{neurovasc} data. To provide visual clarity, only three patient features are shown: age, administration of nimodipine (ATC code: C08CA06), and a tracheostomy procedure (ICD-10-PCS code: 0BH17EZ). (a) Following an initial preprocessing step, (b) the data are passed to MEDS-extract and converted into the MEDS tabular format, where each feature is represented as a distinct event optionally linked to external code systems. (c) In the final stage, the MEDS root folder containing the ETL output is provided as input to the \texttt{meds2rdf} library, which systematically transforms MEDS events into RDF graphs by instantiating the corresponding MEDS-OWL ontology classes. The resulting graphs preserve source information and enable semantic validation through SHACL shapes. 
The exemplified use of our tools to transform \textit{neurovasc} to MEDS and then to RDF is publicly available at \url{https://github.com/albertomarfoglia/meds-to-owl-examples}.

\begin{figure}
  \centering
  \includegraphics[width=1.0\textwidth]{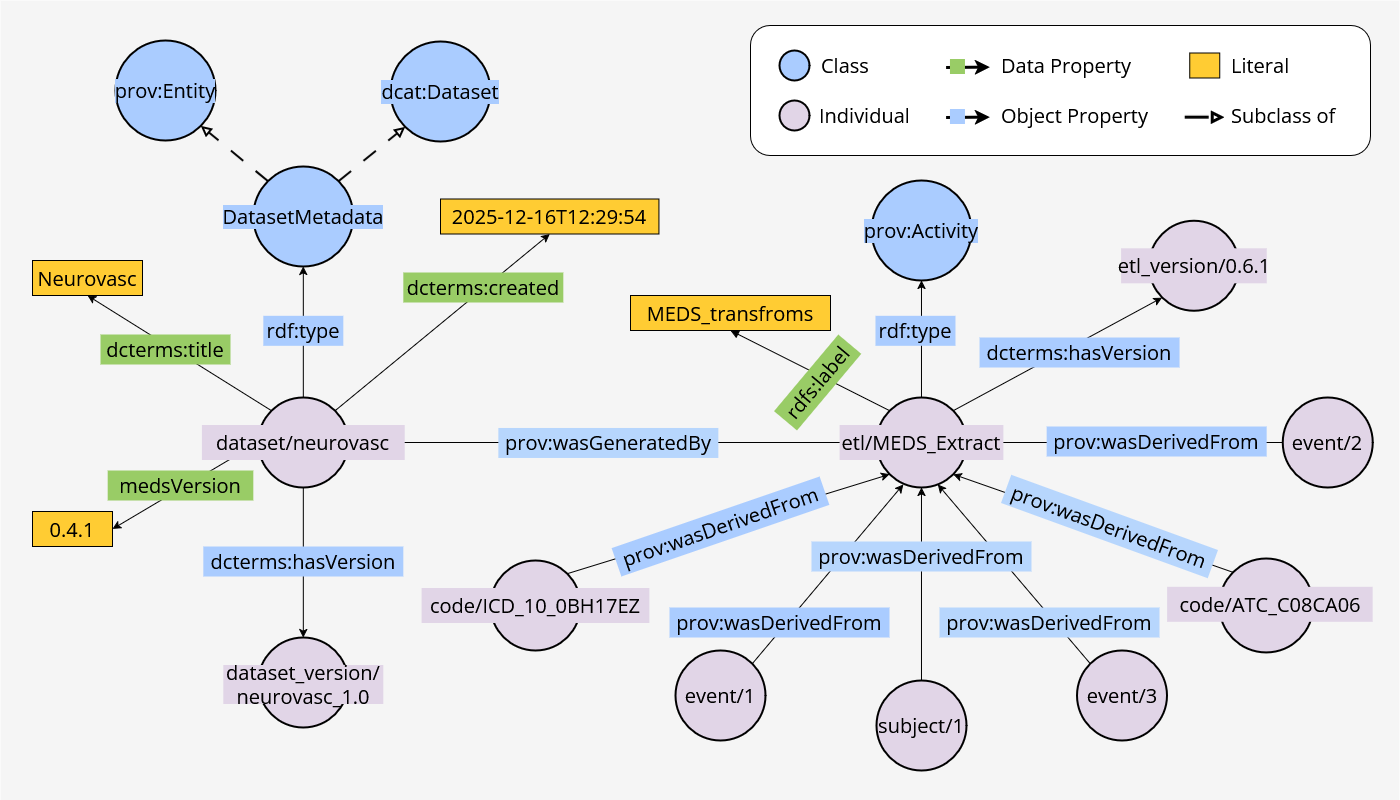}
\caption{Provenance linkage of MEDS events, codes, and subjects using MEDS-OWL, PROV-O, and Dublin Core. Each event references its source \texttt{meds:DatasetMetadata} (here, \textit{neurovasc}) via \texttt{prov:wasDerivedFrom} and is associated with the generating ETL activity (\texttt{prov:Activity}), here MEDS-extract. This approach provides a semantically rich, traceable representation of clinical data workflows.}
  \label{fig:provenance-instance}
\end{figure}

Figure~\ref{fig:provenance-instance} further illustrates how the events shown in Figure~\ref{fig:etl-conversion} are linked to provenance metadata: each event references the originating \texttt{meds:DatasetMetadata} instance via \texttt{prov:wasDerivedFrom}, while the dataset metadata captures creation time, MEDS schema version, dataset title, and is associated with the ETL activity (\texttt{prov:Activity}) that generated it, in this case MEDS-extract. By integrating PROV-O and Dublin Core concepts with MEDS-OWL, this approach provides a semantically rich and traceable representation of clinical data and the associated processing workflow.

To provide a more objective characterization of the transformation process and the resulting KG, we report quantitative statistics at each stage of the \textit{neurovasc} pipeline shown in Figure~\ref{fig:etl-conversion}. The original dataset contains 10,000 rows, each describing 30 patient-level features. After preprocessing and conversion with MEDS-extract, the dataset is transformed into 245,015 MEDS events, reflecting the expansion of patient-level features into event-based representations. These events are distributed across the standard ML splits as follows: 196,064 events in the train set, 24,474 in the validation set, and 24,477 in the test set.

The subsequent conversion of MEDS events into RDF using the \texttt{meds2rdf} library results in a KG comprising 1,330,351 RDF triples. The graph contains 255,135 distinct subjects, 18 predicates, and 46,722 distinct objects, corresponding to 255,161 distinct IRIs and 36,594 literals, with no blank nodes generated. In terms of instantiated ontology elements, the graph includes 245,015 individuals of class \texttt{meds:Event}, corresponding one-to-one with the MEDS events, as well as 10,000 \texttt{meds:Subject} instances and 115 \texttt{meds:Code} instances reused across events. Dataset-level metadata is represented by a single \texttt{meds:DatasetMetadata} individual, while three predefined \texttt{meds:SubjectSplit} individuals capture the train, validation, and test partitions. No label instances are generated for this dataset, as outcome labels are not represented in the current transformation.

Finally, we analyzed the structural footprint of events in the resulting RDF graph. Each \texttt{meds:Event} node is described, on average, by 5.31 RDF triples (standard deviation 0.46), with a minimum of 5 and a maximum of 6 triples per event. This low variance reflects the regular structure imposed by MEDS-OWL cardinality constraints, where subject and code associations are mandatory, while temporal and value-related properties are optional. These statistics demonstrate that the transformation produces a compact RDF representation of events while preserving semantic richness and provenance information.

\textbf{MIMIC-IV example of usage}. To further demonstrate the scalability and applicability of \texttt{meds2rdf} to real-world EHRs, we applied the library to a subset of the publicly available MIMIC-IV demo dataset~\cite{johnson_mimic-iv_2023} already formatted in MEDS~\cite{mcdermott_mimic-iv_2025}. MIMIC-IV is a large, openly accessible critical care dataset containing de-identified EHRs for ICU patients from Beth Israel Deaconess Medical Center. It is widely used in clinical informatics research for benchmarking predictive models, phenotyping, and analyzing temporal data. For this experiment, we selected a subset of 100 patients to illustrate the transformation process while keeping the workflow tractable for visualization and reproducibility.

The conversion generated 916,166 MEDS events distributed across the ML splits as follows: 803,992 in the train set, 69,620 in the tuning set, and 42,554 in the held-out set. The resulting KG comprises 6,408,529 RDF triples, including 925,706 distinct subjects, 19 predicates, 126,680 distinct objects, 925,733 distinct IRIs, and 117,132 literals, with no blank nodes generated. Ontology instantiation includes 916,166 individuals of class \texttt{meds:Event}, 100 \texttt{meds:Subject} individuals, 9,435 \texttt{meds:Code} individuals, three \texttt{meds:SubjectSplit} individuals, and one \texttt{meds:DatasetMetadata} individual. No label or \texttt{ValueModality} instances were included in this subset.  

Each \texttt{meds:Event} in the MIMIC-IV subset is represented by an average of 6.97 RDF triples (median 7.0, minimum 5, maximum 8, standard deviation 0.42), indicating a slightly richer structure per event compared to \textit{neurovasc}, due to the higher number of numeric and text values associated with MIMIC events. These results confirm that \texttt{meds2rdf} can handle real-world MEDS datasets efficiently while preserving the semantic and provenance structure defined by MEDS-OWL.

\section{Discussion}\label{sec:discussion}
%OLD: Our testing on the \textit{neurovasc} dataset shows that MEDS-OWL can capture the MEDS event and their provenance, including their conversion to RDF itself, while remaining compact and pragmatic. The RDF graph produced is SHACL-validated, supporting semantic and structural consistency.
%NEW:
Our evaluation on two complementary datasets, the synthetic \textit{neurovasc} cohort and a real-world subset of MIMIC-IV, demonstrates that MEDS-OWL can faithfully represent MEDS event data and associated provenance in RDF while remaining compact and operationally pragmatic. The resulting RDF graphs are SHACL-validated, supporting semantic and structural consistency.
Moreover, exporting MEDS datasets as KGs enables the direct application of graph representation learning techniques, thereby opening new opportunities for downstream applications such as graph-based ML.
%NEW:
At the same time, the KG representation allows us to leverage the strengths of both paradigms: the event-centric, ML-oriented design of MEDS and the semantic expressivity of KGs. By explicitly modeling entities, relations, and provenance, MEDS-OWL provides semantic enrichment of the underlying datasets, making complex, multi-relational dependencies between clinical events accessible to downstream prediction models. Importantly, because MEDS-OWL preserves the temporal structure and ordering of events, it naturally supports the construction of Temporal Knowledge Graphs (TKGs), enabling longitudinal modeling of patient trajectories and opening new directions for temporally aware clinical prediction tasks.

Compared to existing approaches for representing clinical data in RDF, such as FHIR-RDF, SPHN, and CARE-SM, MEDS-OWL addresses several gaps. Unlike FHIR-RDF, which suffers from loosely specified semantics, or SPHN and CARE-SM, which are tightly coupled to specific initiatives or aims, MEDS-OWL follows the minimality of MEDS, its ML-oriented design, and a well-defined OWL semantics reusing ontologies. This facilitates interoperability across heterogeneous datasets through explicit mapping points (e.g., \texttt{meds:parentCode} or \texttt{skos:exactMatch}).

%OLD: We acknowledge several limitations of the current study. First, the preliminary testing we achieved is limited to the transformation of a single dataset. Transforming other datasets to MEDS-OWL, computing associated stats, and conducting an error analysis of  SHACL inconsistencies potentially arising would enable to validate further the generalizability of our approach. Indeed, we did evaluated the reasoning performance or scalability of the generated KGs; processing large, real-world datasets may reveal bottlenecks in RDF generation, triple-store ingestion, or reasoning that will require optimization.
%NEW:
The dual evaluation highlights both strengths and current limitations. While the \textit{neurovasc} dataset illustrates a test on a controlled synthetic cohort, the MIMIC-IV conversion demonstrates that the approach scales to large, heterogeneous, real-world data. The latter confirms that \texttt{meds2rdf} can process high event volumes while preserving semantic consistency. It also emphasizes the need for careful performance profiling and analysis to avoid SHACL inconsistencies when handling millions of triples and thousands of distinct clinical codes. Currently, we have not conducted an in-depth analysis of reasoning performance or triple-store ingestion costs, which remain important directions for future work in large-scale research applications.
Moreover, the current MEDS-OWL ontology does not capture certain operational constraints of MEDS datasets. Specifically, MEDS requires subject contiguity, which means that all the features associated with a single subject should appear in a single sort and should be temporally sorted. While temporal ordering becomes largely irrelevant when converting tabular MEDS data to RDF graphs, subject contiguity can be partially represented by associating each subject with a specific ML partition via the \texttt{meds:SubjectSplit} class. Importantly, a single subject may participate in multiple independent splits (e.g., for repeated experiments or cross-validation); however, within each split, the subject can belong to only one subset (e.g., training, validation, or test). This ensures that partitioning is mutually exclusive at the level of each experiment while still supporting multiple, independent splits for variability analysis. To address this limitation, future releases of MEDS-OWL will incorporate enhanced axioms, potentially using disjointness rules or qualified cardinality restrictions in description logics, thereby allowing the ontology to more accurately represent MEDS operational constraints.

To substantiate the claimed benefits and broaden the applicability of the ontology, we established an evaluation agenda and identified four steps. 
(1) Quantifying round-trip fidelity to measure information preservation and loss due to transformations. 
(2) Aggregating SHACL validation statistics across multiple, heterogeneous datasets to characterize common data-quality issues and to demonstrate how schema validation reduces downstream errors. 
% OLD:(3) Profile scalability and performance on large, real-world datasets. Specifically, we plan to apply our framework to the MIMIC-IV dataset~\cite{mcdermott_mimic-iv_2025}. Compared to the synthetic \textit{neurovasc} dataset, MIMIC-IV is substantially larger and more complex, providing an opportunity to identify bottlenecks, in addition to assess the semantic fidelity and the scalability of the graph generation process.
% NEW:
(3) Profiling performance on large MEDS corpora will facilitate the identification of bottlenecks and enable a rigorous assessment of semantic fidelity and graph generation scalability.
(4) Compare predictive analysis on MEDS-OWL data with graph-based ML models, such as GCNs. 

Planned extensions of the MEDS-OWL ontology and associated tooling include greater reuse of external ontologies to replace remaining MEDS-specific constructs, thereby reducing bespoke terms and enhancing interoperability. We aim to integrate temporal ontologies (e.g., OWL-Time) and pattern-based representations to more accurately model intervals and episode boundaries, while also adopting unit ontologies to standardize numeric values and support unit conversion and aggregation. Hierarchical relations among codes, currently captured via \texttt{meds:parentCode} (see Table~\ref{tab:codes}), will be formalized using \texttt{skos:broader} to facilitate cross-vocabulary linking, with optional use of \texttt{rdfs:subClassOf} for implementations that model codes as OWL classes. These enhancements aim to increase the semantic richness and usability of MEDS-OWL for KG, enabling systematic cross-vocabulary linking.

\section{Conclusion}\label{sec:conclusion}
We have presented MEDS-OWL, an ontology that formalizes the MEDS event model using Semantic Web standards, along with the \texttt{meds2rdf} library for converting MEDS datasets into RDF graphs compliant with MEDS-OWL. Our framework preserves MEDS’s ML–oriented representation of medical event sequences while enhancing it with a rich semantic structure and explicit links to biomedical and provenance ontologies. This combination enables symbolic reasoning, interoperability across heterogeneous data sources, and seamless integration with knowledge-based systems.

Preliminary experiments on the synthetic \textit{neurovasc} dataset and a real-world subset of MIMIC-IV demonstrate that MEDS-OWL and \texttt{meds2rdf} can accurately transform MEDS events and their associated metadata into semantically validated RDF graphs. Released under an open-source license, these tools promote FAIR-aligned sharing and reuse of medical event data.

By providing a reliable transformation from MEDS to RDF, our framework lays the groundwork for downstream prediction tasks using graph-based ML models.
%NEW:
Representing MEDS datasets as KGs enables semantic enrichment of clinical events and their use across multiple analytical tasks. Moreover, because MEDS-OWL preserves temporal ordering, it naturally supports the construction of TKGs, opening new avenues for longitudinal patient modeling and temporally aware prediction tasks that go beyond conventional tabular approaches.
%

%OLD: In future work, we plan to investigate how RDF-based representations affect predictive performance and to further explore the potential of graph representation learning approaches.
%NEW:
In future work, we plan to systematically compare graph-based ML models with classical ML approaches to assess whether RDF-based, relational representations provide measurable improvements in predictive performance over traditional tabular methods. Additionally, we aim to evaluate the round-trip fidelity of the framework on other public biomedical datasets formatted in MEDS, thereby assessing information preservation across transformations. Finally, we plan to further align MEDS-OWL with widely adopted biomedical ontologies to strengthen semantic interoperability and facilitate broader reuse across biomedical KGs ecosystems.

\section*{Reproducibility Statement}
All experiments and dataset conversions presented in the accompanying paper are fully reproducible using the workflows provided in this repository: \url{https://github.com/albertomarfoglia/meds-to-owl-examples}.
The end-to-end pipelines for the \textit{neurovasc} synthetic dataset and the MIMIC-IV Demo MEDS dataset are implemented as executable, notebook-based workflows (main.ipynb). They rely only on open-source tools and publicly available data or scripts.
The repository specifies all software dependencies and configuration files. Moreover, randomized steps, such as synthetic data generation or sampling, are executed with fixed seeds to ensure repeatability.

\section*{Acknowledgments}
This work is supported by CombO (Health Data Hub) project and the Agence Nationale de la Recherche under the France 2030 program, reference ANR-22-PESN-0007 ShareFAIR, and ANR-22-PESN-0008 NEUROVASC.

\section*{Declaration on Generative AI}
During the preparation of this work, the author(s) used ChatGPT, Grammarly in order to: Grammar and spelling check, Paraphrase and reword. After using this tool/service, the author(s) reviewed and edited the content as needed and take(s) full responsibility for the publication’s content.

% \bibliographystyle{plain}
% \bibliography{main}
\printbibliography

\end{document}